# Learning while Competing - 3D Modeling & Design

Kalind Karia, Rucmenya Bessariya, Krishna Lala, Kavi Arya
Department of Computer Science & Engineering
Indian Institute of Technology Bombay, Powai
Mumbai - 400076, India

*Abstract*—The e-Yantra project at IIT Bombay conducts an online competition, e-Yantra Robotics Competition (eYRC) which uses a Project Based Learning (PBL) methodology to train students to implement a robotics project in a step-by-step manner over a five month period. Participation is absolutely free. The competition provides all resources - robot, accessories, and a problem statement - to a participating team. If selected for the finals, e-Yantra pays for them to come to the finals at IIT Bombay. This makes the competition accessible to resource-poor student teams. In this paper we describe the methodology used in the 6th edition of eYRC, eYRC-2017 where we experimented with a Theme (projects abstracted into rulebooks) involving an advanced topic - 3D Designing and interfacing with sensors and actuators. We demonstrate that the learning outcomes are consistent with our previous studies [1]. We infer that even 3D designing to create a working model can be effectively learnt in a competition mode through PBL.

## I. INTRODUCTION

Competition is an effective way for students to learn topics in embedded systems and robotics. e-Yantra uses Project Based Learning (PBL) methodology in eYRC where students learn while competing and compete while learning [1]. We have discussed the eYRC model in detail [2] and emphasized the need for such hands-on learning in promulgating engineering education to a large number of students across the country at the undergraduate level [3].

Every edition of eYRC comprises several Themes, each developed to solve an abstraction of a real-world problem, such as urban services, smart delivery services and space exploration [4]. Newer technologies are introduced as Themes with increasing complexity for enhanced learning. In last edition of the competition, eYRC-2017, we introduced seven Themes that mimic real-world problems in the agricultural domain for participants to devise innovative solutions, using technology and their creativity.

One such Theme in eYRC-2017, entitled **Spotter Snake**, required students to build a biomorphic hyper-redundant Snake Robot able to traverse uneven terrains and indicate the presence of rodents. These rodents are modeled as colored patches printed on aflex arena. This Theme was unique in that it introduced 3D Designing and 3D Printing to students for the first time. This entailed building a robot by 3D designing its parts and simulating the bot using a free software, printing the parts and later assembling them and integrating the mechatronic components to make the robot perform its tasks.

This required making changes to our eYRC engagement model and the PBL methodology [2] used in previous eYRC editions:
1) Selected teams had at least one student member from a mechanical and one from an electronics discipline.
2) Multiple teams from the same college were allowed to participate if they qualified.
3) The competition was conducted in two stages. Teams that demonstrated good design skills in Stage 1 were selected for Stage 2 and were provided support for 3D printing the robot parts.

## II. MOTIVATION AND RELATED WORK

Inspired by the movement of a snake in an agricultural field, it was decided to design a snake robot to mimic the biological movements of the real snake. The theme entailed a snake moving in the field to "catch" a rodent (i.e. sense a colored marker). This review [5] [6] describes present research efforts related to modeling, analysis and control of snake robots. We used 3D printing of robot parts to design and fabricate a model snake with the right movements. 3D printing is an additive manufacturing process of joining materials to make objects by adding materials layer upon layer. 3D printing hence may be used to make sophisticated custom made designs and allows designers and engineers to create unique products [7].

## III. COMPETITION INSIGHTS

The Spotter Snake Theme was conducted in two stages as illustrated in Fig. 1. Stage 1 consisted of two tasks, viz., Setup, Learn and Design (Task 0), Design and Simulate (Task 1). Task 1 consisted of two sub-tasks, viz., Design 3D Brackets (Task 1.1) and Model and Simulate (Task 1.2).

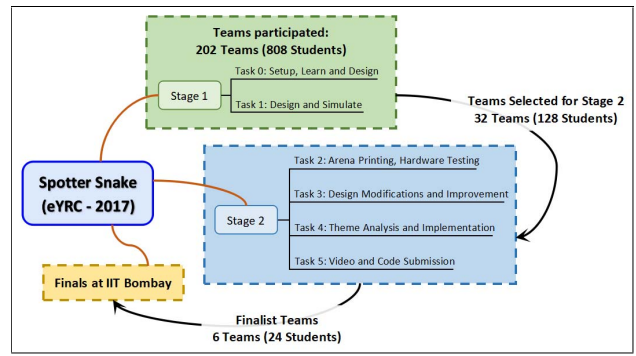

Fig. 1. Spotter Snake Theme Format and Statistics.

### A. Stage 1

*Task 0: Setup, Learn and Design*

Autodesk Fusion 360 (3D designing software) and V-REP (Virtual Robotics Experimentation Platform) were installed and video tutorial [8] provided for learning, practicing designing the enclosure. The 3D model [9] and dimensions of Arduino Uno board [10] were provided for reference. Fig. 2 depicts the enclosure design taught in the tutorial. Designing a similar enclosure for Arduino Nano board was the next task for which the 3D model of the board [11] and mechanical drawing was provided. Table I shows marks distribution for Task 0.

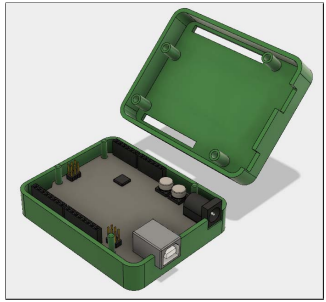

Fig. 2. 3D model of enclosure for Arduino Uno board.

TABLE I
TASK 0 MARKS DISTRIBUTION

| Task 0: Design Arduino Nano board enclosure Total Marks: 50 | |
|---|---|
| *Parameters* | *Marks* |
| Design characteristics, accessibility of pins, port and form factor after 3D printing | 30 |
| Dimensions of design | 20 |

*Task 1: Design and Simulate*

This task was divided into two sub-tasks: Task 1.1 and Task 1.2.

*Task 1.1: Design 3D Brackets*

By understanding and analyzing snakes various gaits and movement dynamics [12] [13] [14] in this task, students designed inter-linkable servo brackets so as to replicate the movement of a snake. This was a key design component to enabling possible "gaits". Resources on various types of mechanical joints and their degrees of freedom were provided. This helped in innovative design-thinking for the brackets of the snake. A 3D model of a servo motor [15] and their mechanical drawings were provided for designing the brackets. Printing material, its density and screw dimensions were also specified.

For this task, following constraints were imposed that are essential for learning product design whilst being mindful of resources:

1) Maximum 3D printing volume of 300 c.c.
2) Number of servo motors - 6 to 10
3) Entire robot shall pass through 100 mm dia. cylinder

Video tutorial [16] was provided to visualize motion study of their bracket design in Autodesk Fusion 360. Teams submitted bracket designs along with the motion study and their Team ID engraved on it on portal. Table II shows marks distribution for Task 1.1.

TABLE II
TASK 1.1 MARKS DISTRIBUTION

| Task 1.1: Design 3D Brackets Total Marks: 60 | |
|---|---|
| *Parameters* | *Marks* |
| Work-ability of joints, inter-linking mechanism and novelty in bracket design | 30 |
| Strength, dimensions, 3D printing volume consumed and engraving of Team ID | 30 |
| If bracket design was found plagiarised more than 20 - 30 % | Disqualified |

*Task 1.2: Model and Simulate*

After designing the brackets, teams had to model and simulate their complete snake robot in V-REP scene which comprised of uneven terrain as a challenging maneuver for the snake robot. Video tutorials on modeling and simulating in V-REP [17] [18] [19] were provided for this task. Teams recorded the V-REP simulation video and submitted the link on portal. Table III shows marks distribution for Task 1.2. Cumulative marks of Task 1.1 and Task 1.2 were considered for selection in Stage 2. Fig. 3 shows marks distribution graph for Stage 1 of 202 teams. Out of these 202 teams, 32 qualified for Stage 2.

TABLE III
TASK 1.2 MARKS DISTRIBUTION

| Task 1.2: Model and Simulate the Snake Robot in V-REP Total Marks: 40 | |
|---|---|
| *Parameters* | *Marks* |
| Maneuvering uneven terrains in each lane avoiding all the obstacles | 20 |
| Bonus if snake performs different gait walks and no penalty is incurred | 20 |

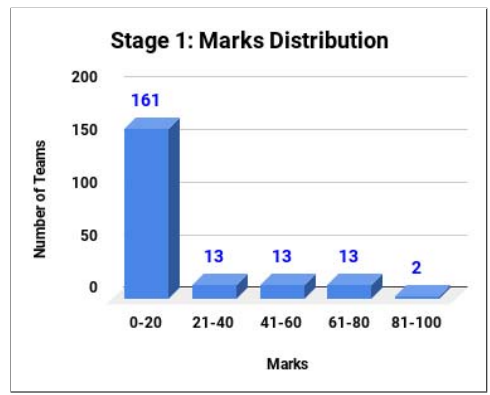

Fig. 3. Stage 1 marks distribution graph.

## B. Stage 2

SLS (Selective Laser Sintering) technique was used for 3D printing of snake robot parts. These parts along with necessary hardware components were shipped to the teams.

The arena of this theme is an abstraction of a godown with food grains stored into silos, that is infested with rodents as shown in Fig. 4. Whilst traversing the arena, the robot had to detect the color and indicate the presence of Rodents by glowing a corresponding colored LED and beeping a buzzer, avoiding the Silos and Rodent Heads.

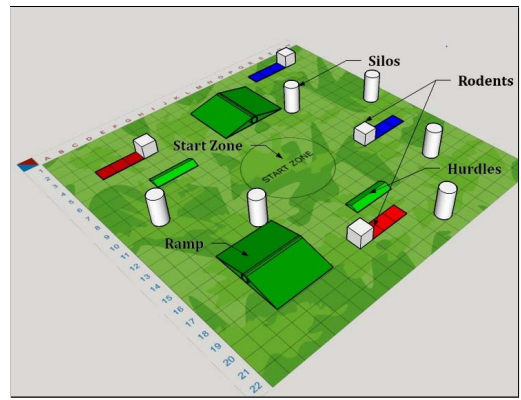

Fig. 4. Spotter Snake theme arena setup.

TABLE IV
MAPPING LEVEL OF LEARNING OUTCOMES TO TASKS AND STATISTICS

| Level* | Level Description* | Task | Learning Imparted | Statistics | |
|---|---|---|---|---|---|
| | | | | Teams (Submitting Task) | Teams (Completed Task) |
| Basic Knowledge | Recognition and understanding of facts, terms, definitions, etc. | Stage 1: Task 0: Setup, Learn and Design  Task 1.1: Design 3D brackets | • Use Autodesk Fusion 360 to 3D design Snake bot parts and basics of V-REP  • Design brackets with limited resource to maximize robot's ability to perform different gaits | 202 | 198 |
| Application of Knowledge | Use of knowledge in ways that demonstrate understanding of concepts, their proper use, and limitations of their applicability | Stage 1: Task 1.2: Model and Simulate  Stage 2: Task 3: Design Modifications and Improvement | • To validate design, its robustness for maneuverability on uneven terrains  • Redesign 3D printed of Stage 1 based on performance and feedback provided | 152 | 32 |
| Critical Analysis | Examination and evaluation of information as required to judge its value to a solution and to make decisions | Stage 2: Task 4: Theme Analysis and and Implementation | • Develop algorithm for snake robot motion to mimic biological movement of natural snake  • Design and fabricate joystick for manually directing the movement of snake robot | 32 | 23 |
| Extension of Knowledge | Extending knowledge beyond what was received, creating new knowledge, making new inferences transferring knowledge to usefulness in new areas of applications | Stage 2: Task 5: Video and Code Submission | • Develop final prototype of snake robot whose movements are directed and controlled by joystick  • Completing the robot's task of detecting all rodents present in the shortest possible time | 23 | 6 |

*Levels and Description of levels are taken from [21]

*Task 2: Arena Printing, Hardware Testing*

Teams were required to print the given arena on a flex sheet and set it up as per the Rulebook. Hardware testing of received components was done for both Transmitter and Receiver system. The connections and assembly for both systems with a sample video [20] were provided.

*Task 3: Design Modifications and Improvement*

Along with marks, feedback was provided to teams during Stage 1 result to modify their bracket designs for optimal usage of allocated printing volume. At this stage, the printing volume was increased to 400 c.c. The final 3D printed bracket parts were shipped to the teams.

*Task 4: Theme Analysis and Implementation*

Teams were quizzed on the information provided in the Rulebook. An example configuration of the arena setup was provided for practicing the solution to the Themes problem statement. This task carried 40 marks.

*Task 5: Video and Code Submission*

In this final task of the competition, sample configurations were provided on which teams submitted a video of their fully implemented solution. This was accompanied by a submission of well commented code. This task carried 50 marks (30 for video and 20 for code). Cumulative marks of Task 4 and Task 5 were considered for selection in Finals. Fig. 5 shows marks distribution for Stage 2 of 32 teams.

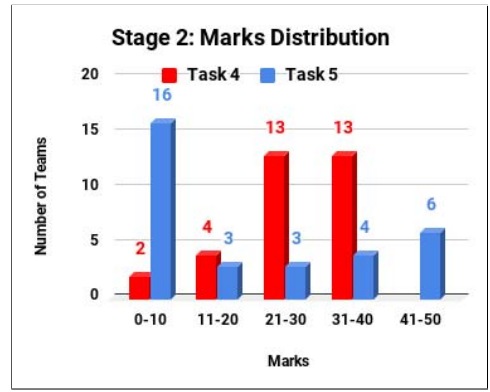

Fig. 5. Stage 2 marks distribution graph.

## C. Finals

The teams that demonstrated the best run (based on the formula given by us in the Rulebook) for given configurations were chosen as Finalists to demonstrate in person at the Finals held at IIT Bombay.

## IV. ANALYSIS OF IMPACT AND EFFECTIVENESS

The primary objective of the e-Yantra competition is learning. In this case, learning online the complex skills related to designing, modeling, simulating, building a complex 3D robot and then using it to solve a given problem. The learning is imparted in a step-by-step manner through Project Based Learning (PBL). Teams of four students each from 681 undergraduate colleges registered amounting to a total of 5932 teams (23728 students) from all across the country. A selection process tested their basic knowledge in electronics, programming and aptitude to qualify to participate. Of these registered teams, 1417 teams qualified and were selected to take part in the competition. These teams were distributed amongst 7 themes (of which the Spotter Snake was one) based on specific criteria.

We analyzed the effectiveness in imparting knowledge in this subject and the learning outcomes in various Tasks assigned for both the Stages and are mapped as represented in Table IV. In this Table, we have shown that in the selected 202 teams, there is learning happening at each level. As illustrated, all 202 teams completed Task 0 while 198 teams completed Task 1. Thus, 98% of teams gained at least **basic knowledge** of 3D designing and modeling. 152 teams out of the 198 teams completed all tasks assigned in Stage 1 - i.e., 77% of these teams were exposed to all the tasks in Stage 1: designing 3D brackets, modeling and simulating in V-REP. Out of these 152 teams, 32 were selected for participation in Stage 2 of the competition - i.e., 21% of these teams could **apply their knowledge** to solve problems.

Out of the 32 teams selected for participation in Stage 2, all the teams completed Task 3 of redesigning their brackets for last phase of 3D printing, which did not carry any marks. 28 teams completed all the tasks in Stage 2: setting up the arena, answering theoretical questions related to assembly of the snake robot with good stability and form factor. These 28 teams applied their knowledge to attempt the tasks, while 23 of them could complete the tasks satisfactorily - 71.8% of these teams demonstrated **critical analysis** skills. Out of these 23 teams, 6 teams showcased their expertise in **extending their knowledge** through a video demonstration of their working solution.

These 6 teams were chosen as finalists and were invited to IIT Bombay to participate in the Finals. Students from the winning teams were assessed for their contribution to the theme and their learning through a viva by a team of judges to get selected for the main prize which is a 6-week residential summer internship at e-Yantra. This ensured that only the deserving candidates from a team of four students were awarded. Five such students from Spotter Snake theme were selected and mentored in the summer internship program for projects like Fish robot, CNC for GrowBox, 3D path planning of drone that utilized their learning and knowledge gained from participating in the competition.

## V. CONCLUSION

We have demonstrated that complex training in designing 3D printed artifacts was possible remotely while the students were participating in a competition. Training was imparted on an important and versatile technology such as 3D designing, modeling and simulation and electronic integration of a complex articulated snake robot. Even though 23 out of 202 teams (11%) reached the final stage of building and animating the snake it may be noted that 32 teams (16%) were able to reach the point of mastering the tools to design, simulate and build a snake robot. A large percentage of the participants do get training in Stage 1 and are unable to qualify for Stage 2 as they find it difficult to balance their academic curriculum and the competition which is a good educational exercise but not a part of their curriculum. With this limitation, we believe this to be a respectable success ratio for imparting complex skills to an inexperienced cohort of students in a challenging time of 5 months from start to finish.

The goal of the e-Yantra project is to rapidly train large numbers of engineering students in practical skills through its various initiatives, the e-Yantra National Robotics Competition (eYRC) competition being just one of them. The purpose is "learning to compete whilst competing to learn". We have witnessed through our competition that students who would previously never have encountered robotics have, through the e-Yantra competition, discovered a newfound enthusiasm for engineering and for robotics in particular. This training is highly empowering for students and leads them onto the path of becoming potential innovators which is a serious requirement for a growing economy such as India.

## ACKNOWLEDGMENT

We thank Dr. Saraswathi Krithivasan, for her valuable inputs in devising the Theme problem statement and storyline. We are grateful to MHRD for funding e-Yantra project. We acknowledge support of e-Yantra project staff especially, Abhinav Sarkar, Saurav Shandilya, Sachin Patil, Aditya Panwar, Piyush Manavar, Sanam Shakya and Lohit Penubaku.